\documentclass[conference]{IEEEtran}

\usepackage{cite}
\usepackage{amsmath,amssymb,amsfonts}
\usepackage{algorithmic}
\usepackage{graphicx}
\usepackage{textcomp}
\usepackage{xcolor}

\usepackage{multirow}  
\usepackage{subcaption}  
\usepackage[normalem]{ulem}  
\usepackage{subfiles}  
\usepackage{booktabs}
\usepackage{url}
\usepackage{amsthm}

\newcommand{\ie}{\textit{i}.\textit{e}., }

\useunder{\uline}{\ul}{}

\makeatletter 
\newcommand{\linebreakand}{%
  \end{@IEEEauthorhalign}
  \hfill\mbox{}\par
  \mbox{}\hfill\begin{@IEEEauthorhalign}
}
\makeatother 

\newtheorem{theorem}{Theorem}[section]

\newtheorem{lemma}[theorem]{Lemma}
\newtheorem{definition}{Definition}[section]

\begin{document}


\title{
    On the Relationship between \\
    Populated Regions and Adversarial Robustness \\
    in Deep Neural Networks
}


\author{
    \IEEEauthorblockN{Seongjin Park\textsuperscript{\textasteriskcentered}}
    \IEEEauthorblockA{
        \textit{Samsung AI Center} \\
        Suwon, South Korea \\
        ssjin.park@samsung.com
    }
\and
    \IEEEauthorblockN{Haedong Jeong\textsuperscript{\textasteriskcentered}}
    \IEEEauthorblockA{
        \textit{Sogang University} \\
        Seoul, South Korea \\
        haedong@sogang.ac.kr
    }
\and
    \IEEEauthorblockN{Tair Djanibekov\textsuperscript{\textasteriskcentered}}
    \IEEEauthorblockA{
        \textit{MBZUAI} \\
        Abu Dhabi, United Arab Emirates \\
        tair.djanibekov@mbzuai.ac.ae
    }
\linebreakand
    \IEEEauthorblockN{Giyoung Jeon}
    \IEEEauthorblockA{
        \textit{LG AI Research} \\
        Seoul, South Korea \\
        giyoung.jeon@lgresearch.ai
    }
\and
    \IEEEauthorblockN{Jinseok Seol\textsuperscript{\textdagger}}
    \IEEEauthorblockA{
        \textit{Dankook University} \\
        Yongin, South Korea \\
        jinseok.seol@dankook.ac.kr
    }
\and
    \IEEEauthorblockN{Jaesik Choi\textsuperscript{\textdagger}}
    \IEEEauthorblockA{
        \textit{KAIST \& INEEJI} \\
        Daejeon \& Seongnam, South Korea \\
        jaesik.choi@kaist.ac.kr
    }
}


\maketitle

\begingroup
    \renewcommand
    \thefootnote{\textasteriskcentered}
    \footnotetext{ equally contributed authors}
\endgroup
\begingroup
    \renewcommand
    \thefootnote{\textdagger}
    \footnotetext{ both are corresponding authors}
\endgroup


\begin{abstract}

    In general, deep neural networks (DNNs) are evaluated by the generalization performance measured on unseen data excluded from the training phase.
    Along with the development of DNNs, the generalization performance converges to the state-of-the-art performances and it becomes difficult to evaluate DNNs solely based on this metric.
    The robustness against adversarial attack has been used as an additional metric to evaluate DNNs by measuring their vulnerability.
    However, few studies have been performed to analyze the adversarial robustness in terms of the geometry in DNNs.
    In this work, we perform an empirical study to analyze the internal properties of DNNs that affect model robustness under adversarial attacks.
    In particular, we propose the novel concept of the populated region set (PRS), where training samples are actually populated, to represent the internal properties of DNNs in a practical setting.
    From systematic experiments with the proposed concept, we provide empirical evidence to validate that a low PRS ratio has a strong relationship with the adversarial robustness of DNNs.
    We also devise a PRS regularizer leveraging the characteristics of PRS to improve the adversarial robustness without adversarial training.

    
\end{abstract}




\section{Introduction}

    \begin{figure}[tp]
    \centering
    \includegraphics[width=1.0\linewidth]{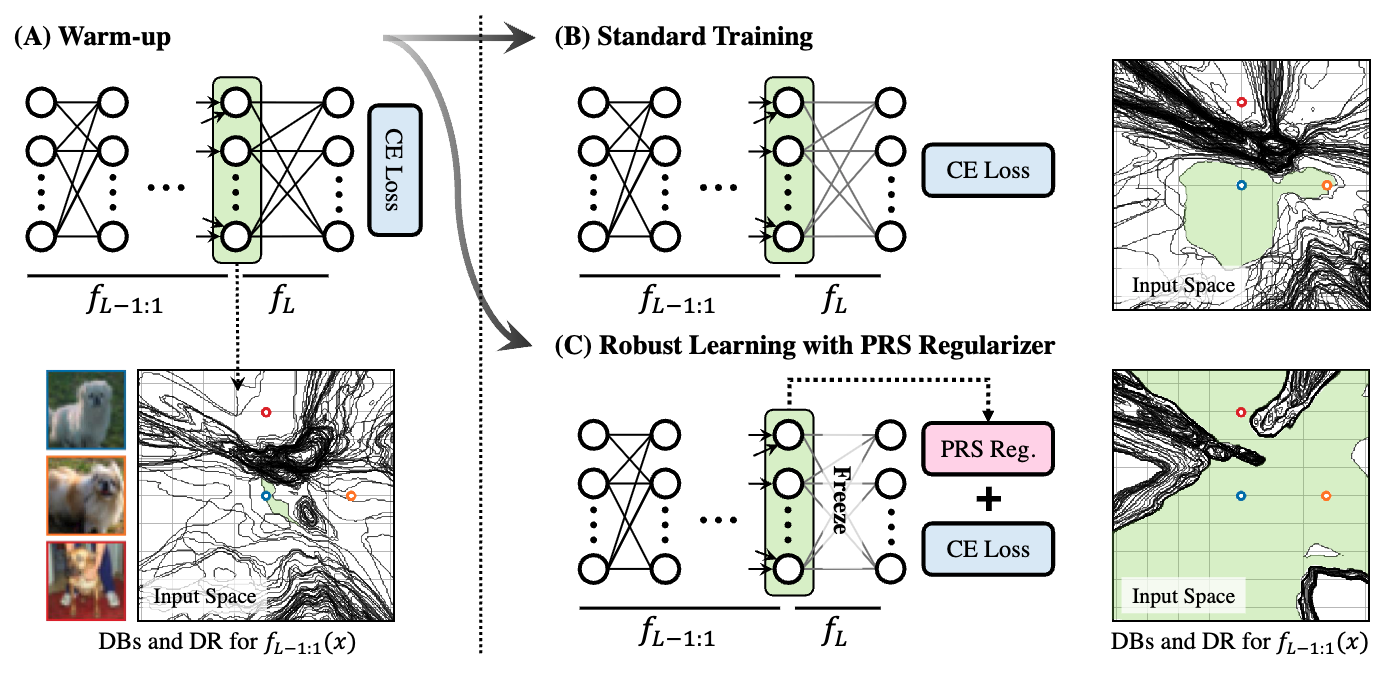}
    \caption{
        An illustrative comparison of each training method with CIFAR-10, and visualization for decision boundaries/regions (DBs/DRs) of penultimate layer in the input space ($f^{(L-1:1)}(x)$).
        For visualization, we randomly select three dog images and depict section of input space.
        The green area indicates DR which the blue boxed image populates.
        (A) Warm-up stage for VGG-16 with standard training (cross-entropy loss).
        (B) Standard training after warm up stage.
        (C) The robust learning with devised PRS regularizer after warm up stage.
        We identify that each training scheme induces different configuration of DBs/DRs, which represents different internal properties of DNNs.
    }
    \label{fig:o1}
\end{figure}

    With the steep improvement of the performance of Deep Neural Networks (DNNs), their applications are expanding in the real world \cite{o17}.
    For real world application, it may be necessary to choose the best model among the candidates.
    Traditionally, the generalization performance which measures the objective score on the test dataset excluded in the training phase, is used to evaluate the models \cite{o2}.
    However, it is non-trivial to evaluate DNNs based on this single metric.
    For example, if two networks with the same structure have the similar test accuracy, it is ambiguous which is better.
    Robustness against adversarial attacks, measure of the vulnerability, can be an alternative to evaluate DNNs \cite{o9, o10, o12, o14, o26, o31, o32}.
    Adversarial attacks aim to induce model misprediction by perturbing the input with small magnitude.
    Most previous works were focused on the way to find adversarial samples by utilizing the model properties such as gradients with respect to the loss function.
    Given that the adversarial attack seeks to find the perturbation path on the model prediction surface over the input space, robustness can be expressed in terms of the geometry of the model.
    However, few studies have been performed to interpret the robustness with the concept of the geometric properties of DNNs.
    From a geometric viewpoint, the internal properties of DNNs are represented by the boundaries and the regions.
    It is shown that the DNNs with piece-wise linear activation layers are composed of many linear regions, and the maximal number of these regions is mathematically related to the expressivity of DNNs \cite{o22, o28}.
    As these approaches only provide the upper bound for the expressivity with the same structured model, it does not explain how much information the model actually expresses.


\begin{figure}[tp]
    \centering
    \includegraphics[width=1.0\linewidth]{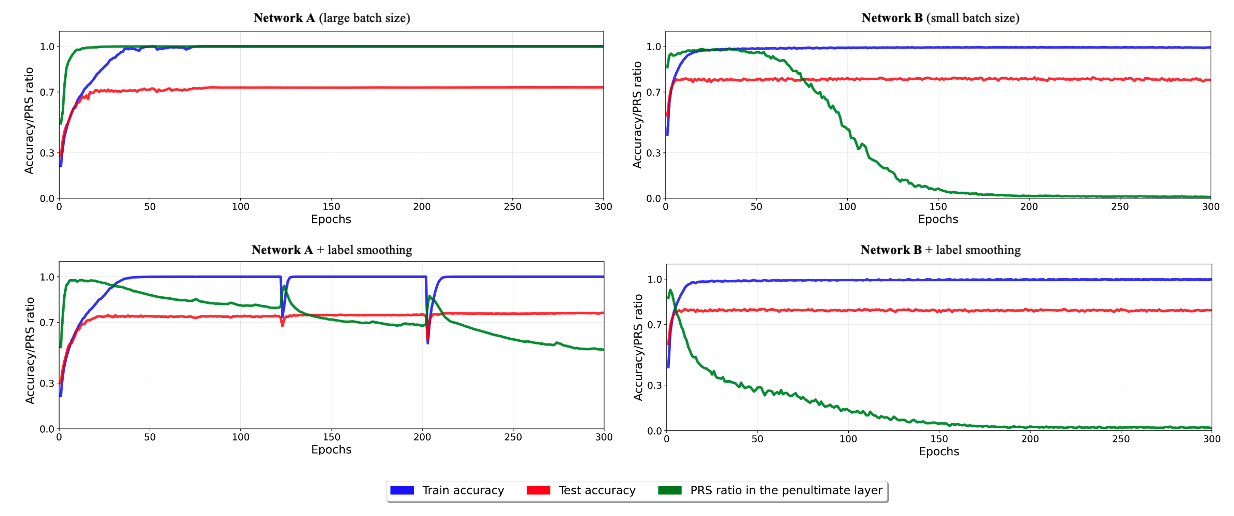}
    \caption{
        Training/test accuracy and the PRS ratio on the penultimate layer on CNN-6 with batch size 2048 and 128.
        We select the networks at the $300^{\text{th}}$ epoch and denote these two CNN-6 by Network A and B, respectively, throughout the paper (PRS ratio of Network A: 0.99, and Network B: 0.007).
        We also should be scenario where the label smoothing is applied.
    }
    \label{fig:o2}
\end{figure}

    In this work, we investigate the relationship between the internal properties of DNNs and the adversarial robustness.
    In particular, our approach analyzes the internal characteristics from the perspective of the decision boundary (DB) and the decision region (DR), which are basic components of DNNs \cite{o7}.
    To avoid insensitivity of the maximal number of linear regions in the same structure assumption, we propose the novel concept of the populated region set (PRS), which is a set of DRs containing at least one sample included in the training dataset.
    Since the PRS can be considered as the feasible complexity of the model, we hypothesize that the size of PRS is related to the robustness of network.
    To validate our hypothesis, we perform systematic experiments with various structures of DNNs and datasets.
    Our observations are summarized as follows:

    \begin{enumerate}
        \item
            The models with the same structure can have different size of PRS, although they have similar generalization performance.
            We empirically show that the model with a small size of the PRS tends to show higher robustness compared to that with a large size.
            (in Section \ref{sec:prs-and-robustness})
        \item
            We observe that when the model achieves a low PRS ratio, the linear classifier which maps the penultimate features to the logits has high cosine similarity between parameters corresponding to each class (in Section \ref{sec:prs-and-robustness}.
        \item
            We verify that the size of intersection of the PRS from the training/test dataset is related to the robustness of model.
            The model with a high PRS inclusion ratio of test samples has higher robustness than that with a low PRS inclusion ratio (in Section  \ref{sec:prs-and-test}).
        \item
            We devise a novel regularizer leveraging the characteristics of PRS to improve the robust accuracy without adversarial training (in Section  \ref{sec:prs-regularizer}).
        \item
            We also provide a theoretical analysis of the relationship between the proposed PRS ratio and techniques known to enhance robustness, such as adjusting the batch size and applying label smoothing (in Section \ref{sec:theory}). 
    \end{enumerate}


\section{Related Work}
\label{sec:related-work}

    The adversarial attack which reveals the vulnerability of DNNs, is mainly used to validate the reliability of the trained network.
    As an early stage for adversarial attacks, the fast gradient sign method (FGSM) \cite{o9} based on the gradient with respect to the loss function and the multi-step iterative method \cite{o16} are proposed to create adversarial examples to change the model prediction with a small perturbation.
    Recently, many studies on effective attacks in various settings have been performed to understand the undesirable decision of the networks \cite{o3, o20, o24}.
    In terms of factors affecting robustness, \cite{o30} provide evidence to argue that training with a large batch size can degrade the robustness of the model against the adversarial attack from the perspective of the Hessian spectrum.
    
    With increasing interest in the expressive power of DNNs, there have been several attempts to analyze DNNs from a geometric perspective \cite{o4, o6}.
    In these studies, the characteristics of the decision boundary or regions formulated by the DNNs are mainly discussed.
    \cite{o22} show that the cascade of the linear layer and the nonlinear activation organizes the numerous piece-wise linear regions.
    They show that the complexity of the decision boundary is related to the maximal number of these linear regions, which is determined by the depth and the width of the model.
    \cite{o28} extend the notion of the linear region to the convolutional layers and show the better geometric efficiency of the convolutional layers.
    Compared to previous work, our work focuses on the practical decision region which the trained network actually utilizes.
    It has also been shown that the manifolds learned by DNNs and the distributions over them are highly related to the representation capability of a network \cite{o19}.
    While these studies highlight the benefits of increasing expressivity of DNNs as the number of regions increases, interpreting the vulnerability of DNNs with the geometry is another important topic.
    \cite{o29} show that a model with thick decision boundaries induces robustness.
    \cite{o23} show that a decision boundary with a small curvature acquires the high robustness of the model.
    These approaches focus on the decision boundaries, while this paper suggests to focus on the decision regions, which are composed by the surrounding decision boundaries.


\section{Problem Setting}
\label{sec:problem}

    This section describes the internal properties of DNNs from the perspective of DBs and DRs.
    The DBs of the DNN classifier is mainly defined as the borderline between DRs for classification, where the prediction probability of two different classes are the same \cite{o8}.
    To expand the notion of DBs and DRs to the internal feature-level, we re-define the DBs in the classifier that generalizes the existing definition of DBs.
    We then propose the novel concept of the Populated Region Set (PRS) that describes the specific DRs used from the network for training samples.

    \subsection{Decision Boundary and Region}

        Let the classifier with $L$ layers be $F(x) = (f^{(L)} \circ \phi \circ f^{(L-1)} \circ \phi \circ \cdots \phi \circ f^{(1)})(x) = f^{(L:1)}(x)$, where $x$ is the sample in the input space $\mathcal{X} \subset \mathbb{R}^{d_{\mathcal{X}}}$ and $\phi(\cdot)$ denotes the non-linear activation function\footnote{Although there are various activation functions, we only consider ReLU activation for this paper.}.
        For the $l$-th layer, $f^{(l)}(\cdot)$ denotes a linear operation and $f^{(l:1)}_{i}(\cdot)$ denotes the value of the $i$-th element of the feature vector $f^{(l:1)}(x) \in \mathbb{R}^{d_{l}}$.
        We define the decision boundary for the $i$-th neuron of the $l$-th layer.

        \begin{definition}[Decision Boundary (DB)]
            The $i$-th \textbf{decision boundary} $\text{DB}^{(l)}_{i}$ at the $l$-th layer is defined as:
            \begin{equation}
                \text{DB}^{(l)}_{i} = \{x \in \mathcal{X} \mid f^{(l:1)}_{i}(x) = 0\}.
            \end{equation}
        \end{definition}

        \noindent
        We note that the $\text{DB}^{(l)}_{i}$ with $l < L$ divides the input space $\mathcal{X}$ based on the hidden representation of the $l$-th layer (\ie existence of feature and the amount of feature activation).
        There are $d_{l}$ boundaries and the configuration of the $\text{DB}$s are arranged by the training.
        As input samples in the same classification region are considered to belong to the same class, the input samples placed on the same side of the internal $\text{DB}^{(l)}_{i}$ share the similar feature representation.
        In this sense, we define the decision region, which is surrounded by $\text{DB}$s.

        \begin{definition}[Decision Region (DR)]
            Let $\sigma \in \{-1, +1\}^{d_{l}}$ be a signature vector.
            Then the \textbf{decision region} $\text{DR}^{(l)}_{\sigma}$, which shares the signature of the feature represenation, is defines as:
            \begin{equation}
                \text{DR}^{(l)}_{\sigma} = \{
                    x \in \mathcal{X}
                    \mid
                    \text{sign}(f_{l:1}(x)) = \sigma
                \}.
            \end{equation}
        \end{definition}

        Fig. \ref{fig:o1} presents each training scheme for VGG-16 with CIFAR-10 and the internal DBs/DRs of the penultimate layer ($f^{(L-1:1)}(x)$).
        To visualize the DBs and DRs in the 2D space, we randomly select three training images (red, blue, and orange box) in dog class and make a hyperplane with these images.
        The standard training ((A) $\rightarrow$ (B)) and the proposed robust learning ((A) $\rightarrow$ (C)) are performed after warm-up stage (a stage without the regularizer), respectively.
        We identify that the proposed regularizer induces different configuration of DBs/DRs in the input space compared to the standard training ((A) $\rightarrow$ (B)).


    \subsection{Populated Region Set}
    \label{sec:problem-setting-prs}

        It is well-studied that the number of DRs is related to the representation power of DNNs \cite{o22, o28}.
        In particular, the expressivity of DNNs with partial linear activation function is quantified by the maximal number of the linear regions and this number is related to the width and depth of the structure.
        We believe that although the maximal number can be one measure of expressivity, the trained DNNs with finite training data\footnote{In general, the number of training data is smaller than the maximal number of the linear region.} cannot handle the entire regions to solve the task.
        To only consider DRs that the network uses in the training process, we devise the train-related regions where training samples are populated more frequently.
        We define the populated region set (PRS), which is a set of DRs containing at least one sample included in the training dataset.
        PRS will be used to analyze the relationship between the geometrical property and the robustness of DNNs in a practical aspect.

        \begin{definition}[Populated Region Set (PRS)]
            From the set of every DRs of the model $f$ and given the dataset $\mathcal{D}$, the \textbf{populated region set} $\text{PRS}^{(l)}$ is defined as:
            \begin{equation}
                \text{PRS}^{(l)}(\mathcal{D}) = \{
                    \text{DR}^{(l)}_{\sigma}
                    \mid
                    \text{DR}^{(l)}_{\sigma} \cap \mathcal{X}_{\mathcal{D}} \neq \emptyset
                \},
            \end{equation}
            where $\mathcal{X}_{\mathcal{D}}$ denote the inputs of the dataset.
        \end{definition}

        \noindent
        We note that the size of the $\text{PRS}$ is bounded to the size of given dataset $\mathcal{D}$.
        When $|\text{PRS}^{(l)}| = \mathcal{D}$, each sample in training dataset is assigned to each distinct DR in the $l$-th layer.
        To compare the PRS of networks, we define the PRS ratio as $|\text{PRS}^{(l)}|/|\mathcal{D}|$, which measures the ratio between the size of the PRS and the given dataset.
        Fig. \ref{fig:o2} presents a comparison between two equivalent neural networks (A and B) with six convolution blocks (CNN-6) trained on CIFAR-10 varying only the batch size (2048 and 128, respectively).
        Here, we first observe the following:
        (1) Training with a large batch size is known to be less robust compared to its small-batch counterpart from various perspectives, and we can see that this phenomenon is expressed through the PRS ratio.
        Going forward, we will utilize Networks A and B, which differ only in PRS, to make direct comparisons.
        (2) Conversely, label smoothing is known to enhance robustness, and we confirm that this effect also manifests itself through the PRS ratio.
        We observe that even with larger batch sizes, the PRS ratio tends to decrease gradually.
        Inspired by this observation, we incorporate aspects of label smoothing when designing a PRS-based regularizer later.
        The theoretical analysis of these two observations (batch size and label smoothing in relation to the PRS ratio) will be discussed in Section \ref{sec:theory}.
        From the fact that the penultimate layers are widely used as feature extraction, we only consider the PRS ratio on the penultimate layer in the remainder of the paper.


\section{Robustness Under Adversarial Attacks}

    \begin{figure}[tp]
    \centering
    \includegraphics[width=0.8\linewidth]{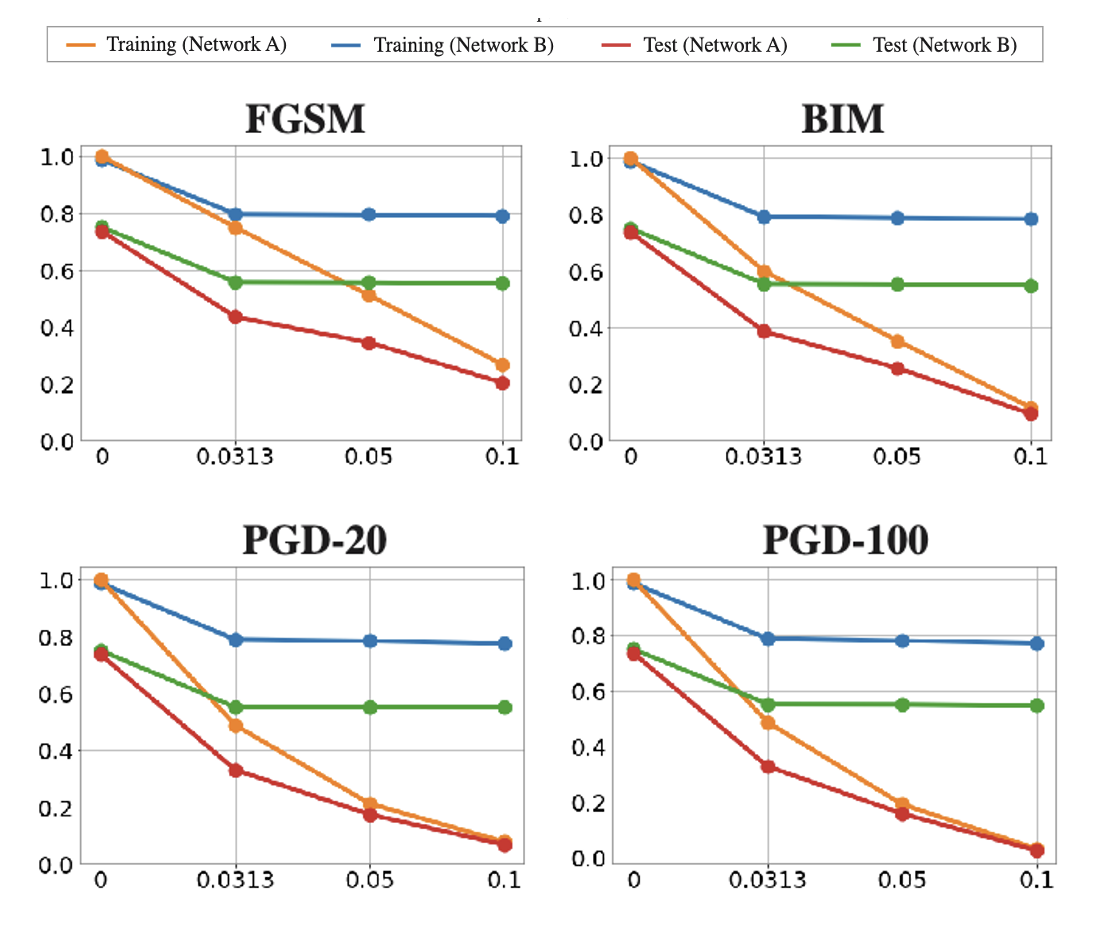}
    \caption{
        Robust accuracy under various adversarial attack methods on networks A and B.
        The x-axis indicates perturbation $\epsilon$ and the y-axis indicates the training/test robust accuracy.
    }
    \label{fig:o3}
\end{figure}

    \begin{figure*}[tp]
    \centering
    \includegraphics[width=0.8\linewidth]{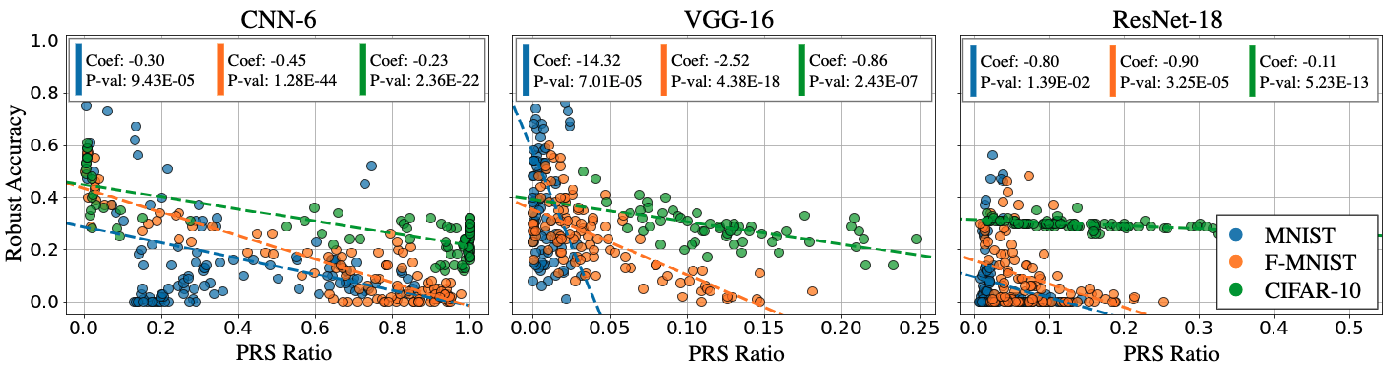}
    \caption{
        Relationship between the PRS ratio and robust accuracy attacked by PGD method in various models and datasets.
        The colored dots are for the independent models.
        The colored dashed lines indicate the trend for each dataset.
    }
    \label{fig:o4}
\end{figure*}

    In this section, we perform experiments to analyze the relationship between the PRS ratio and the robustness.
    We evaluate the robustness of the network using the fast gradient sign method (FGSM) \cite{o9}, basic iterative method (BIM) \cite{o16} and projected gradient descent (PGD) \cite{o20} method widely used for the adversarial attacks.
    The untargeted adversarial attacks using training/test dataset are performed for the various perturbations ($\epsilon = 0.0313, 0.05, 0.1$).


    \subsection{Experimental Setup}

        For the systematic experiments, we select three different structures of DNNs to analyze:
        (1) a convolutional neural network with six convolution blocks (CNN-6),
        (2) VGG-16 \cite{o25}, and
        (3) ResNet-18 \cite{o11}.
        We train\footnote{Cross-entropy loss and Adam optimizer with learning rate $10^{-3}$ is used.} basic models with fixed five random seeds and four batch sizes (64, 128, 512 and 2048) over three datasets: MNIST, F-MNIST \cite{o27}, and CIFAR-10 \cite{o15}.
        For the extensive analysis on the correlation between the PRS ratio and properties of network, we extract candidates from each basic model with the various epochs.
        Then we apply the test accuracy threshold to guarantee the sufficient and similar performance.


    \subsection{PRS and Robustness}
    \label{sec:prs-and-robustness}

        First, we compare the two models (Network A and B in Fig. \ref{fig:o2}) with similar test accuracy but different PRS ratio\footnote{We note that different PRS ratios are obtained by different batch size of Network A (2048) and B (128).}.
        Fig. \ref{fig:o3} presents the results of robust accuracy (accuracy against adversarial attack) under the FGSM, BIM (5-step), PGD-20 (20-step), and PGD-100 (100-step) on $L_{\infty}$ with $\alpha = 2/255$.
        We identify that Network B (low PRS ratio) is more robust than Network A (high PRS ratio) under all adversarial attacks.

        As the PGD-20 shows the similar robust accuracy compare to a PGD-100, we focus on an analysis under the PGD-20 in the rest of the paper.
        We measure the PRS ratio and the robust accuracy in all models and datasets to verify the relationship between the PRS ratio and the robustness.
        For the experiments, we take the magnitude of $\epsilon$ as follow: MNIST = 0.3, F-MNIST = 0.1, and CIFAR10 = 0.0313 on $L_{\infty}$ norm.
        Fig. \ref{fig:o4} presents the experimental results according to the model structure under the PGD attack.
        To quantify the relation, we calculate the coefficient of the regression line and perform significance test to validate the trend.
        From Fig. \ref{fig:o4}, we identify that the PRS ratio has an inversely correlated relationship with the robust accuracy in most cases.

        From the above observations, we empirically confirm that the PRS ratio is
        related to the robustness against adversarial attacks.
        In order to investigate the evidence that the low PRS ratio causes robustness for the gradient-based attack, we perform an additional analysis of failed attack samples.
        In the gradient-based attack, as the magnitude of the gradient is a crucial component to success, we first count the ratio of the zero gradient samples in the failed attack samples.

\begin{figure*}
    \centering
    \hfill
    \begin{subfigure}[t]{0.23\linewidth}
        \includegraphics[width=\linewidth]{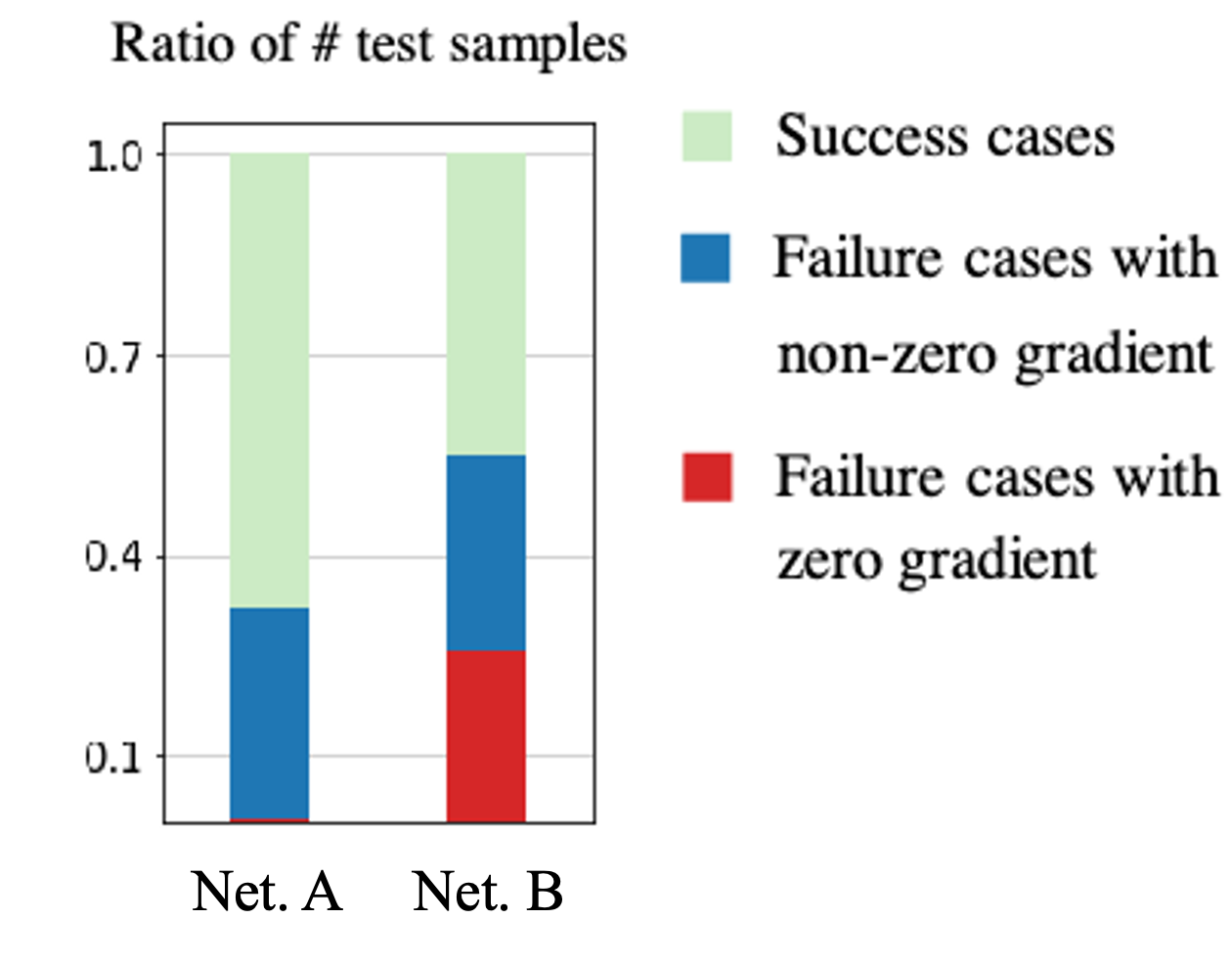}
        \caption{Ratio of test samples}
        \label{fig:o5-a}
    \end{subfigure}
    \hfill
    \begin{subfigure}[t]{0.65\linewidth}
        \includegraphics[width=\linewidth]{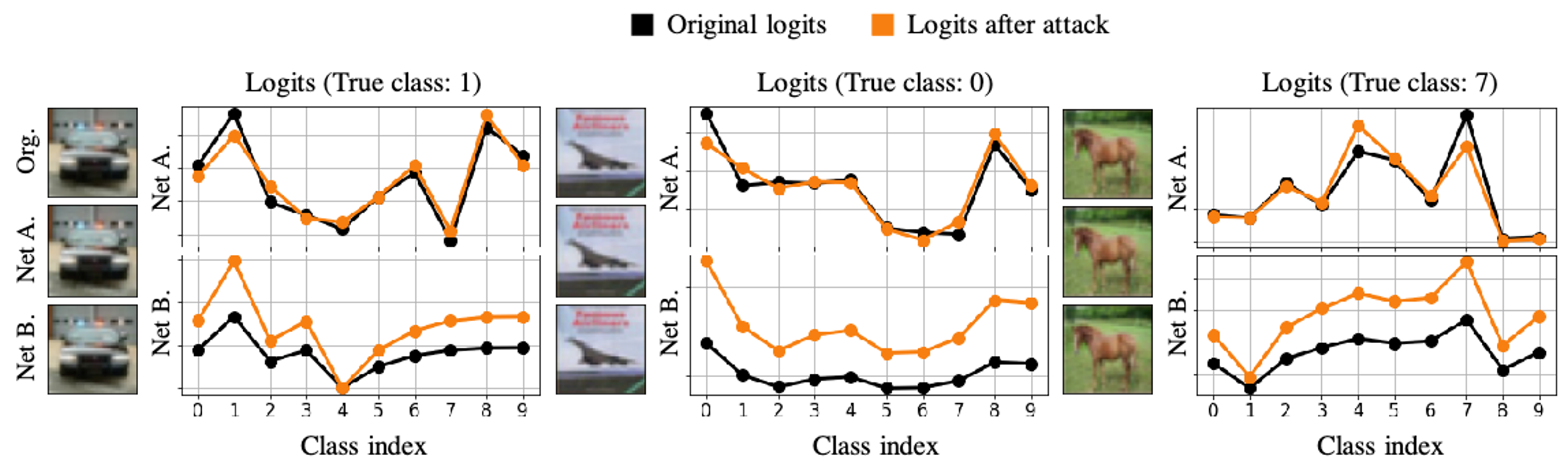}
        \caption{Illustrative examples for change of logits in attack failure cases}
        \label{fig:o5-b}
    \end{subfigure}
    \hfill
    \caption{
        (a)
            Comparison of the ratio of the zero gradient in the failure attack for the test samples under the PGD-20 attack on $L_{\infty}$ with $\epsilon = 0.0313$ (Network A and B).
        (b)
            The illustrative examples of attacked samples on Network A and B which is failed on B, and the corresponding logits before/after the attack.
            After the attack, the logits move on almost parallel direction with the original logits in Network B.
    }
    \label{fig:o5}
\end{figure*}

        Fig. \ref{fig:o5-a} shows the ratio of success samples (light green bar), failure samples with non-zero gradient (blue bar) and zero gradient (red bar) in all samples.
        We note that the failed attack samples with non-zero gradients maintain the index
        of the largest logit as the true class after attack.
        To analyze the reason of failure, we examine the change of the logits under the adversarial attack.
        This change is shown in Fig. \ref{fig:o5-b}.
        To clarify the difference of the change of the logits between Network A and B, we select the examples of successful attack on Network A but failed attack on Network B.
        In Network B, the logits move on almost parallel direction, which causes the predicted label to be maintained as the true class.

        \begin{figure}
    \centering
    \hfill
    \begin{subfigure}[t]{0.16\linewidth}
        \includegraphics[width=\linewidth]{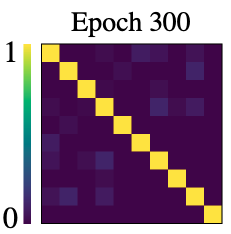}
        \caption{Net. A}
        \label{fig:o6-a}
    \end{subfigure}
    \hfill
    \begin{subfigure}[t]{0.80\linewidth}
        \includegraphics[width=\linewidth]{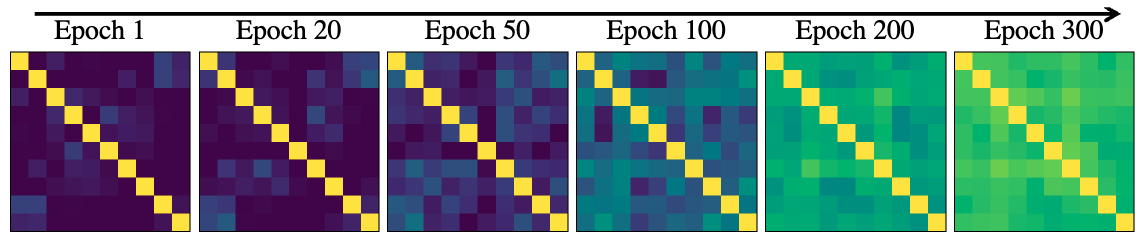}
        \caption{Net. B}
        \label{fig:o6-b}
    \end{subfigure}
    \hfill
    \caption{
        (a)
            Cosine similarity (CS) matrix for a final layer on Network A.
        (b)
            CS matrix for epochs on Network B.
            As the epoch increases, the CS for each parameter increases.
    }
    \label{fig:o6}
\end{figure}

        To explain the parallel change of the logit vector, we hypothesize that the DBs corresponding to each class node have similar configuration in the input space.
        However, it is intractable to measure the similarity between DBs in the entire network due to the highly non-linear structure and the high dimensional input space.
        To simplify our hypothesis, we only measure the cosine similarity between the parameters which map the features on the penultimate layer to logits (\ie final layer).
        Fig. \ref{fig:o6} presents that the similarity matrices for Networks A and B.
        When we compare the matrix between the two models at the 300$^{\text{th}}$ epoch, we identify that Network B (low PRS ratio) has higher cosine similarity between each parameter in the final layer.
        We note that the cosine similarity between each parameter in the final layer can be considered as the degree of parallelism for the normal vectors in the linear classifier.
        We also confirm that the decrease of the PRS ratio is aligned with the increase of the similarity of parameters in Fig. \ref{fig:o6-b}, when we consider the graph in Fig. \ref{fig:o2}.
        To verify the relationship between PRS ratio and the cosine similarity, we measure the PRS ratio and the cosine similarity between each parameter in all models.
        Table \ref{tab:o1} shows the results of the correlation experiment for the relationship between the PRS ratio and the cosine similarity.
        We identify that the PRS ratio has an inverse correlation for the cosine similarity between each parameter in the final layer.


    \subsection{PRS and Test Samples}
    \label{sec:prs-and-test}

        When we regard the model as a mapping function from the input space to the
        feature space, handling unseen data in a known feature domain is significant
        in a perspective of the generalization performance.
        Hence, if the majority of samples from the test dataset are assigned to the training PR, the model can be considered to be learned the informative and general concept of feature mapping.
        For example, if the arbitrary test sample is mapped to the training PR, we
        expect that a similar decision will appear.
        However, it is non-trivial to guess which type of decision will appear when the test sample is mapped to out of the training PR.
        To investigate the differences between the test samples which are included and excluded in the training PR, we evaluate the test accuracy under adversarial attack for each group.
        For a comparison, we divide both the inclusion and exclusion groups with $1k$ correctly predicted test samples.
        
\begin{figure}[tp]
    \centering
    \includegraphics[width=0.75\linewidth]{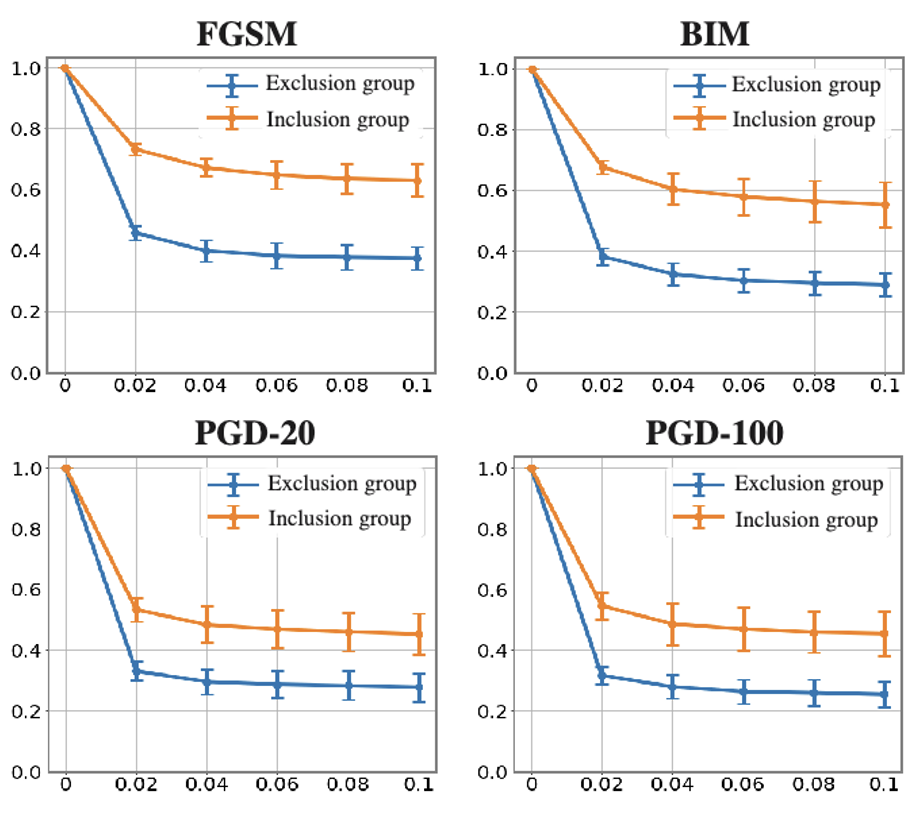}
    \caption{
        Test accuracy under adversarial attacks for inclusion/exclusion groups for CNN6 on CIFAR-10 for five fixed random seeds.
        The x-axis indicates perturbation $\epsilon$ and the y-axis indicates the robust accuracy.
        The blue/orange line indicates the exclusion/inclusion groups, respectively.
        The exclusion group is shown to be more vulnerable under adversarial attacks.
    }
    \label{fig:o7}
\end{figure}

        Fig. \ref{fig:o7} shows the robust accuracy under the FGSM, BIM with a 5-step, and
        the PGD-20 and PGD-100 on $L_{\infty}$.
        Although the robust accuracy of each test group decreases as the epsilon becomes larger, we observe that the inclusion group is more robust against all types of attacks compared to the exclusion group.
        Table \ref{tab:o1} presents the results of the correlation experiment for the
        relationship between the PRS ratio and the inclusion ratio of the test samples for
        the training PR.
        We compute the inclusion ratio as the ratio of the test samples mapped to the training PR.
        In Table \ref{tab:o1}, we identify that the PRS ratio and the inclusion ratio have inversely correlated relationship.
        As we previously verify that the included test samples show high robustness, we empirically confirm that the low PRS ratio is related to the robustness under adversarial attacks.

    \subsection{PRS and Training Samples}
    
        \begin{figure*}
    \centering
    \hfill
    \begin{subfigure}[t]{0.6\linewidth}
        \includegraphics[width=\linewidth]{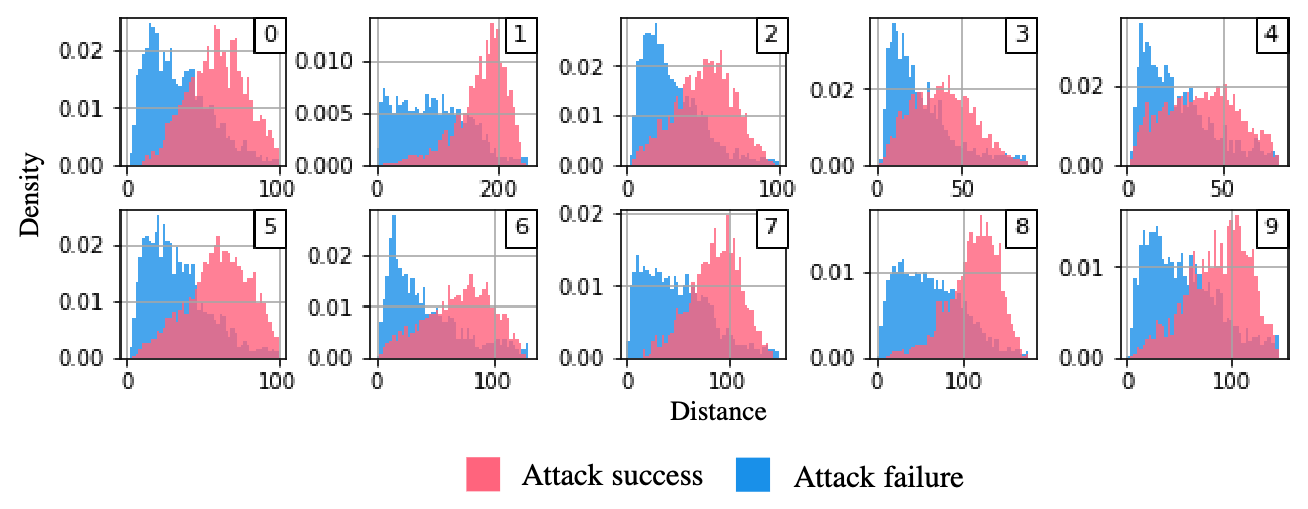}
        \caption{Distribution of the distance between the each feature vector and the MRV}
        \label{fig:o9-a}
    \end{subfigure}
    \hfill
    \begin{subfigure}[t]{0.25\linewidth}
        \includegraphics[width=\linewidth]{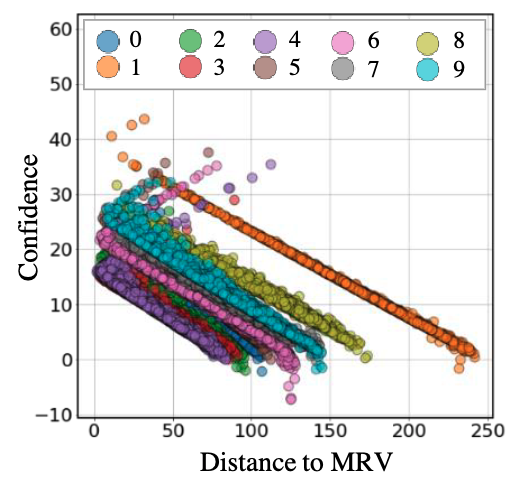}
        \caption{Relationship between the distance to the MRV and the confidence}
        \label{fig:o9-b}
    \end{subfigure}
    \hfill
    \caption{
        Experimental results under VGG-16 on CIFAR-10.
        (a)
            Distribution of Euclidean distance to MRV for training samples.
            The blue histogram indicates the failed attack samples and the red histogram indicates the success attack samples.
            The white box in the upper right presents each class.
        (b)
            Relationship between the distance to MRV and the confidence.
            The colored dots represent training samples which are vulnerable under adversarial attack per each class.
    }
    \label{fig:o9}
\end{figure*}

        \begin{figure}
    \centering
    \hfill
    \begin{subfigure}[t]{0.32\linewidth}
        \includegraphics[width=\linewidth]{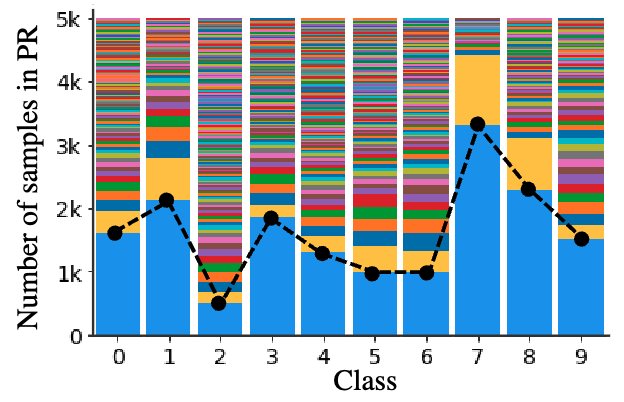}
        \caption{The number of training samples in DR}
        \label{fig:o8-a}
    \end{subfigure}
    \hfill
    \begin{subfigure}[t]{0.295\linewidth}
        \includegraphics[width=\linewidth]{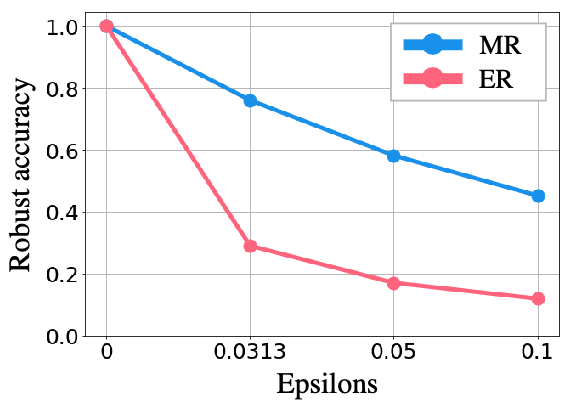}
        \caption{Robust accurcy for MR and ER}
        \label{fig:o8-b}
    \end{subfigure}
    \hfill
    \begin{subfigure}[t]{0.305\linewidth}
        \includegraphics[width=\linewidth]{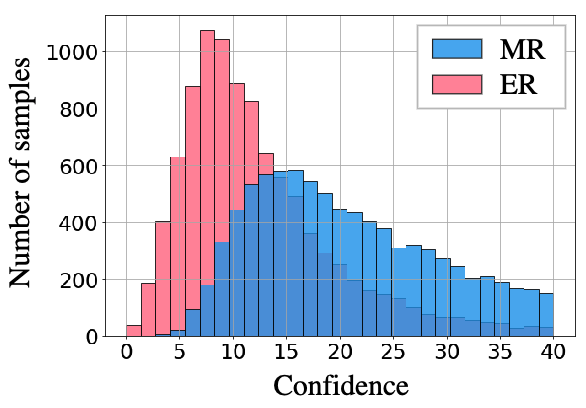}
        \caption{Class confidence for MR and ER}
        \label{fig:o8-c}
    \end{subfigure}
    \hfill
    \caption{
        Experimental results under VGG-16 on CIFAR-10.
        (a)
            The number of training samples populated in DR for each class.
            The maximum number of samples for each class is 5k and the sky blue bar indicates each MR.
        (b)
            Robust accuracy under PGD-20 attacks on $L_{\infty}$ for the samples in MR and ER.
        (c) Class confidence for the samples in MR and ER.
    }
    \label{fig:o8}
\end{figure}

        From previous section, we empirically observe that the vulnerability of individual test samples is related to PRS defined by training samples.
        In this section, we categorize the PRS to expand this relationship.
        At first, we define the major DR for each class c which includes the majority of training samples.

        \begin{definition}[Major Region (MR)]
            For a training dataset $\mathcal{D}_{c}$ denoting the instances with class $c$, the \textbf{major region} for $l$-th layer and class $c$ is defined as:
            \begin{equation}
                \text{MR}^{(l)}_{c} = \underset{\text{DR}^{(l)}_{\sigma} \in \text{PRS}^{(l)}(\mathcal{D}_{c})}{\text{argmax}}
                |\mathcal{X}_{\mathcal{D}_{c}} \cap \text{DR}^{(l)}_{\sigma}|
            \end{equation}
        \end{definition}

        \noindent
        We note that since the training samples are finite, large number of samples occupying $\text{MR}^{(l)}_{c}$ directly means the low PRS ratio.
        We refer the remained DRs (\ie not MR) as the \textbf{extra region}s (ERs).

        \subsubsection{Comparison of MR and ER}
        
            At first, we observe the distribution of training samples for type of region corresponding each class in VGG-16 trained with CIFAR-10.
            Fig. \ref{fig:o8-a} depicts the distribution of training samples for MR (sky
            blue and black dashed line) and ER for each class.
            Although the training samples are distributed the various regions, in almost case, we identify that there are MR for entire class.
            To compare the characteristics of samples populated each region, we randomly selected $10k$ training samples from MR and ER.
            We perform adversarial attack for selected samples and measure the confidence of the prediction (logit value for the target class).
            Fig. \ref{fig:o8-b} and \ref{fig:o8-c} show the robust accuracy and confidence for MR and ER, respectively.
            We empirically verify the training samples in MR have higher adversarial robustness and the network predicts these samples with high confidence.

        \subsubsection{Relationship between MR and Confidence}
        
            From the empirical observations that samples belonging to MR are relatively robust, we hypothesize that samples located closer to center of MR tends to be more robust.

        \begin{definition}[Major Region Mean Vector (MRV)]
            For a major region $\text{MR}^{(l)}_{c}$, the \textbf{major region mean vector} is defined as:
            \begin{equation}
                \text{MRV}^{(l)}_{c} = \frac{1}{|\text{MR}^{(l)}_{c}|} \sum_{x \in \text{MR}^{(l)}_{c}} f_{l:1}(x).
            \end{equation}
        \end{definition}

        \noindent
        To verify our hypothesis, we measure Euclidean distance between MRV and training samples with success/failure of adversarial attack.
        From Fig. \ref{fig:o9-a}, we identify that the samples far from MRV tend to be vulnerable for the adversarial attack in the entire class.
        Furthermore, in Fig. \ref{fig:o9-b}, we identify the inversely correlated relationship between the confidence and the distance to MRV for the failed attack samples.


\section{Robust Learning via PRS Regularizer}
\label{sec:prs-regularizer}

    In previous Sections, we empirically verify that PRS is related to the adversarial robustness.
    In particular, (1) inclusion of MR, and (2) distance to MRV are highly related to vulnerability of individual samples.
    From these insights, we devise a novel regularizer leveraging the properties of PRS to improve the adversarial robustness.


    \subsection{Regularizer via PRS}

        At first, we design the regularizer to reduce the distance between feature vector
        and MRV.
        To guarantee the quality of feature representation which constructs plausible PRS, we utilize the warm-up stage for the classifier.
        In the warm-up stage, the classifier is trained with cross-entropy loss function $\mathcal{L}_{\text{ce}}$ during the $T$-th epoch.
        After warm-up stage, we construct the MRV for each class and use it after $T$-th epoch.
        The regularizer for MRV is defined as:

        \begin{equation}
            \mathcal{L}_{\text{MRV}} = \frac{1}{|\mathcal{D}|} \sum_{(x, y) \in \mathcal{D}} \left(\text{MRV}^{(l)}_{y} - f^{(l:1)}(x)\right)^{2}.
        \end{equation}

        \noindent
        We note that because $\mathcal{L}_{\text{MRV}}$ reduce the distance to MRV in the feature space, the arbitrary sample can have the opportunity for inclusion of MR.
        However, it is non-trivial to guarantee for inclusion based on Euclidean distance, because (1) in general, the feature vector is encoded in the high dimensional space, and (2) highly non-linear embedding of DNNs.


        As mentioned in Section \ref{sec:problem-setting-prs}, we additionally blend label smoothing techniques to ensure that the training samples are included into MR.
        We adopt label smoothing $\alpha$ into the MR regularizer, whereas we pull a feature into corresponding MRV upto $(1 - \alpha)$ (compared to the original MRV regularizer), while also pulling the feature towards MRV of other classes in the ratio of $\alpha$.
        We denote this regularizer as $\mathcal{L}_{\text{MRV}}^{\alpha}$.

        The final objective becomes:
        
        \begin{equation}
            \mathcal{L}_{\text{PRS}} = \lambda_{\text{ce}} \mathcal{L}_{\text{ce}} + \lambda_{\text{MRV}}\mathcal{L}_{\text{MRV}}^{\alpha}
        \end{equation}

        \noindent
        where $\lambda_{\text{ce}}$ and $\lambda_{\text{MRV}}$ are hyperparameters.
        We perform simple grid search to set hyperparameters and use $\lambda_{1} = 0.2$, $\lambda_{2} = 0.8$, and $\alpha = 0.1$ for remaining experiments.
        We denote the loss function with $\alpha = 0$ by $\mathcal{L}_{\text{MR}}$.


    \subsection{Experimental Results}


\begin{figure}[tp]
    \centering
    \includegraphics[width=0.8\linewidth]{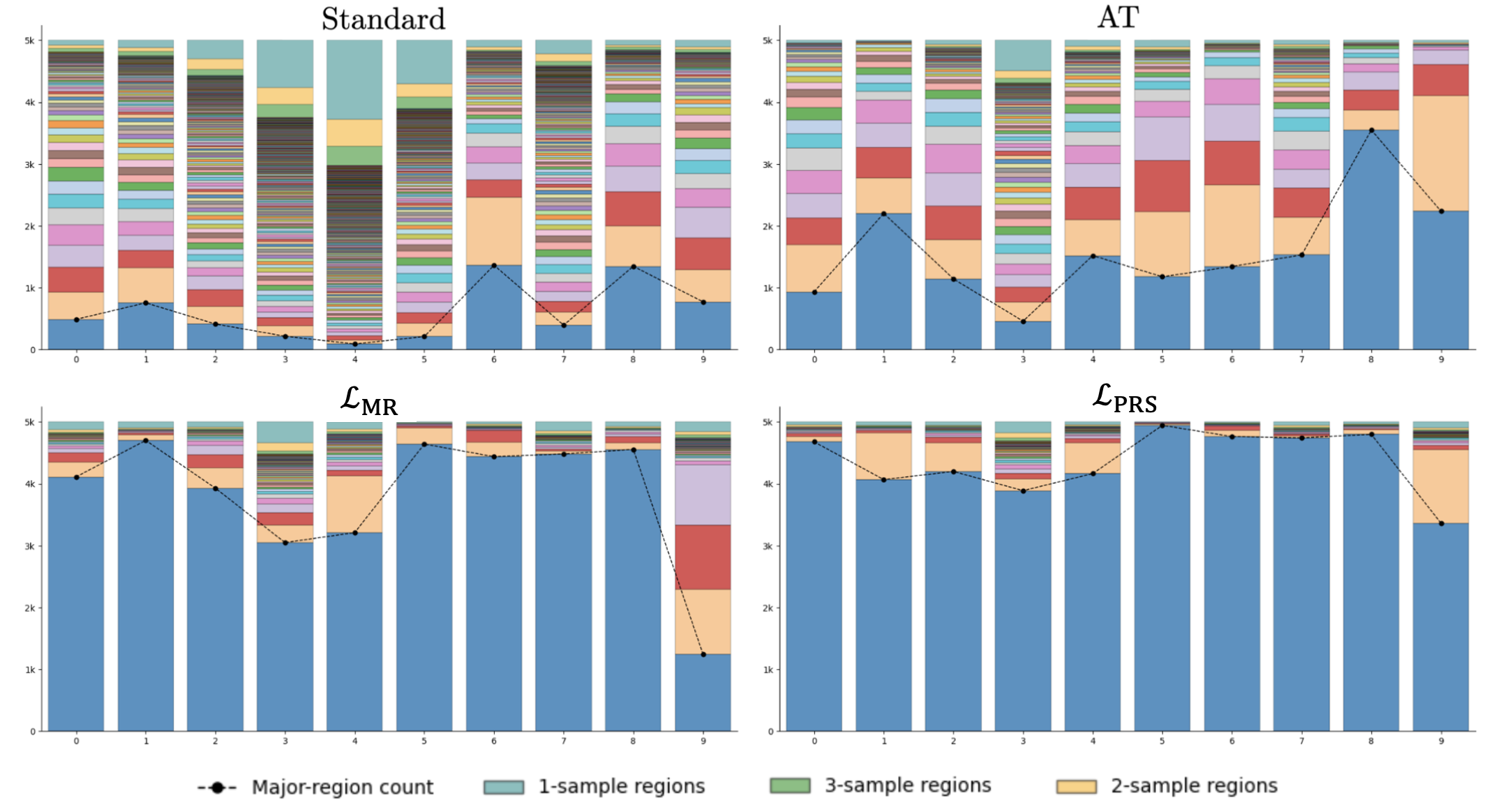}
    \caption{
        Experimental results under VGG-16 on CIFAR-10.
        The distribution of DR for each class by our proposed method.
        The sky blue bar per class indicates each MR.
    }
    \label{fig:o10}
\end{figure}

        To verify the effectiveness of our proposed method, we apply the regularizer to
        CNN-6, VGG-16 and ResNet-18 for the classification task on CIFAR-10.
        We set standard training ($\mathcal{L}_\text{ce}$) and Adversarial training (AT) based on PGD-20 attack on $L_{\infty}$ as the baselines.
        For warm-up stage, we use parameters of classifier with standard training ($T = 50$), and freeze the final layer parameters to observe the effect of the change of PRS for robust and test accuracy.
        Fig. \ref{fig:o1} depicts the training procedure for each training scheme and trained DBs and DRs which represent the state of PRS.
        Table \ref{tab:o2} shows the results for each training scheme on various architectures on CIFAR-10.
        We verify that the proposed method can improve the robust accuracy while maintaining the test accuracy.
        We note that the proposed PRS regularizer does not use the adversarial examples to improve the adversarial robustness.
        It means that our method can have strength in the perspective of computation time.
        We further investigate the change of PRS after training for each method.
        Fig. \ref{fig:o10} shows MR and ER of each class for each training scheme.
        We identify that the both proposed regularizers can increase the number of populated training samples in MR compared to the standard training (\ie reduce of PRS ratio).
        \emph{During the period of reviewing process, we share our source code in a anonymized repository}\footnote{https://anonymous.4open.science/r/PRS\_ICDM25}.

        \begin{table}[tp]
    \centering
    \caption{The coefficient and $p$-values of the regression analysis between the
PRS ratio and each properties.}
    \label{tab:o1}
    \begin{tabular}{c|c|rc|rc}
        \toprule
            \multirow{2}{*}{Model} &
            \multirow{2}{*}{Dataset} &
            \multicolumn{2}{c|}{Cosine Similarity} &
            \multicolumn{2}{c}{Inclusion Ratio} \\ 
            & &
                coef. & $p$-value &
                coef. & $p$-value \\
        \midrule
            \multirow{3}{*}{CNN-6}
                & MNIST & -0.76 & 5.17e-29 & -0.72 & 1.09e-126 \\
                & F-MNIST & -0.58 & 2.43e-46 & -0.79 & 1.78e-133 \\
                & CIFAR-10 & -0.65 & 4.35e-54 & -0.80 & 5.45e-136 \\
        \midrule
            \multirow{3}{*}{VGG-16}
                & MNIST & -14.70 & 2.70e-8 & -0.53 & 4.66e-43 \\
                & F-MNIST & -3.38 & 4.11e-19 & -0.48 & 2.22e-24 \\
                & CIFAR-10 & 0.28 & 2.92e-1 & -0.79 & 2.07e-9 \\
        \midrule
            \multirow{3}{*}{ResNet-18}
                & MNIST & -2.28 & 1.36e-11 & -0.53 & 1.58e-98 \\
                & F-MNIST & -2.24 & 1.66e-16 & -0.65 & 6.38e-38 \\
                & CIFAR-10 & -1.29 & 4.20e-14 & -0.71 & 2.87e-35 \\
        \bottomrule
    \end{tabular}
\end{table}

        \begin{table}[tp]
    \centering
    \caption{Comparison of robust and test accuracy under PGD-20 attacks on $L_{\infty}$ for
CIFAR-10.}
    \label{tab:o2}
    \begin{tabular}{c|ccccr}
        \toprule
            Model & Method & Robust Acc. & Test Acc. & PRS Ratio \\
        \midrule
            \multirow{5}{*}{\rotatebox[origin=c]{90}{\parbox[c]{1cm}{\centering CNN-6}}}
                & Standard
                    & 38.82 $\pm$ 2.73 & 77.92 $\pm$ 0.34 & .101 $\pm$ .010 \\
                & AT
                    & 53.79 $\pm$ 0.42 & 70.65 $\pm$ 0.17 & .099 $\pm$ .001 \\
                & $\mathcal{L}_{\text{MR}}$ (ours)
                    & 53.59 $\pm$ 0.07 & 80.30 $\pm$ 0.72 & .018 $\pm$ .003 \\
                & $\mathcal{L}_{\text{PRS}}$ (ours)
                    & 57.10 $\pm$ 0.45 & 80.47 $\pm$ 0.09 & .009 $\pm$ .002 \\
        \midrule
            \multirow{5}{*}{\rotatebox[origin=c]{90}{\parbox[c]{1cm}{\centering VGG-16}}}
                & Standard
                    & 39.94 $\pm$ 1.28 & 80.28 $\pm$ 0.24 & .115 $\pm$ .012 \\
                & AT
                    & 58.18 $\pm$ 0.13 & 75.22 $\pm$ 0.05 & .069 $\pm$ .002 \\
                & $\mathcal{L}_{\text{MR}}$ (ours)
                    & 60.42 $\pm$ 0.36 & 78.61 $\pm$ 0.15 & .038 $\pm$ .007 \\
                & $\mathcal{L}_{\text{PRS}}$ (ours)
                    & 63.55 $\pm$ 0.72 & 79.31 $\pm$ 0.22 & .018 $\pm$ .003 \\
        \midrule
            \multirow{5}{*}{\rotatebox[origin=c]{90}{\parbox[c]{1cm}{\centering ResNet-18}}}
                & Standard
                    & 33.48 $\pm$ 0.08 & 76.96 $\pm$ 0.15 & .065 $\pm$ .001 \\
                & AT
                    & 50.65 $\pm$ 0.20 & 73.03 $\pm$ 0.05 & .046 $\pm$ .004 \\
                & $\mathcal{L}_{\text{MR}}$ (ours)
                    & 49.31 $\pm$ 0.65 & 76.51 $\pm$ 0.08 & .061 $\pm$ .003 \\
                & $\mathcal{L}_{\text{PRS}}$ (ours)
                    & 50.59 $\pm$ 0.12 & 76.81 $\pm$ 0.11 & .059 $\pm$ .002 \\
        \bottomrule
    \end{tabular}
\end{table}


\section{Theoretical Analysis}
\label{sec:theory}

    In this section, we provide a theoretical analysis about the relationship between the batch size and PRS ratio, and also the relationship between the label smoothing and PRS ratio, whereas those two robustness-related components were heavily affecting the PRS ratio.

\subsection{Preliminaries}

    For concreteness, we define and redefine some concepts with more rigorous notations.

    \begin{definition}[Class Decision Boundary]
        In a softmax-based neural network model for classification task, a \textbf{logit} $l_{c; \theta}$ is defined per class $c$ as follows:
        \begin{equation}
            l_{c; \theta}(x) = w_{c; \theta}^{\top}z_{x; \theta} + b_{c; \theta},
        \end{equation}
        where $w_{c; \theta}$, $b_{c; \theta}$ denote the weight and bias parameter in the last layer for the class $c$, and $z_{x; \theta}$ is features of the penultimate layer for the input $x$.
        A \textbf{class decision value} $\text{CDV}_{c, c'}(x; \theta)$ is defined by the difference between logits of the two classes $c$ and $c'$ for an input $x$:
        \begin{equation}
            \text{CDV}_{c, c'}(x; \theta) = l_{c; \theta}(x) - l_{c'; \theta}(x).
        \end{equation}
        A \textbf{class decision boundary} $\text{CDB}_{c, c'}(\theta)$ is a set of features with zero decision values:
        \begin{equation}
            \text{CDB}_{c, c'}(\theta) = \{z_{x; \theta} \mid \text{CDV}_{c, c'}(x; \theta) = 0\}.
        \end{equation}
    \end{definition}
    
    \begin{definition}[Class Decision Region]
        A \textbf{class decision region} $\text{CDR}_{c}(\theta)$ for class $c$ is defined as a intersection of regions determined by decision boundaries with other classes $c'$ other than $c$:
        \begin{equation}
            \text{CDR}_{c}(\theta) = \bigcap_{c' \neq c}
            \left\{
                z_{x; \theta} \mid \text{CDV}_{c, c'}(x; \theta) > 0
            \right\}
            .
        \end{equation}
    \end{definition}

    \begin{definition}[Class Decision Distance]
        A \textbf{class decision distance} $\text{CDD}$ is the minimal Euclidean distance from the feature point $z_{x; \theta}$ to a decision boundary:
        \begin{equation}
            \text{CDD}(z_{x; \theta}, \text{CDB}_{c, c'}(\theta)) = \frac{|\text{CDV}_{c, c'}(x; \theta)|}{\|w_{c; \theta} - w_{c'; \theta}\|}.
        \end{equation}
    \end{definition}

    \begin{definition}[Margin Distance]
        A \textbf{margin distance} $\text{MD}$ is the minimal CDD:
        \begin{equation}
            \text{MD}(z_{; \theta}) = \min_{c' \neq c} \text{CDD}(z_{x; \theta}, \text{CDB}_{c, c'}(\theta)).
        \end{equation}
    \end{definition}



\subsection{Batch Size and PRS Ratio}

    The proof sketch proceeds as follows:
    Large batch sizes are known to yield sharp minima due to the Hessian eigenvalue argument \cite{keskar2017large}, and such sharp minima induce higher parameter sensitivity.
    Highly sensitive parameters lead to sensitive decision boundaries, resulting in smaller decision regions.
    Ultimately, these smaller decision regions cause a higher PRS ratio.

    \begin{definition}[Sharpness]
        Let $\theta \in \mathbb{R}^{p}$ denote parameters for an empirical loss function $\mathcal{L}$, then for a minima $\theta^{*}$ where $\nabla \mathcal{L}(\theta^{*}) = 0$, we define \textbf{sharpness} $\text{S}_{\mathcal{L}}(\theta^{*})$ by the largest eigenvalue of the Hessian at the minima, namely
            $\text{S}_{\mathcal{L}}(\theta^{*}) = \lambda_{\text{max}} (H(\theta^{*}))$,
        where $\lambda_{\text{max}}$ denote the largest eigenvalue of $H(\theta^{*}) = \nabla_{\theta}^{2} \mathcal{L}(\theta^{*})$, and we say a minima $\theta^{*}$ is sharper then another minima $\theta^{**}$ whenever $\text{S}_{\mathcal{L}}(\theta^{*}) > \text{S}_{\mathcal{L}}(\theta^{**})$.
    \end{definition}

    For a small perturbation $\Delta\theta \in \mathbb{R}^{p}$, let us expand $\mathcal{L}(\theta)$ around a minima $\theta^{*}$ using the second-order Taylor expansion:
    \begin{equation}
        \mathcal{L}(\theta^{*} + \Delta\theta) \approx \mathcal{L}(\theta^{*}) + \nabla_{\theta}\mathcal{L}(\theta^{*})^{\top}\Delta\theta + \frac{1}{2} \Delta\theta^{\top} H(\theta^{*}) \Delta\theta.
    \end{equation}
    
    Note that since $\theta^{*}$ is a minima, the gradient at $\theta^{*}$ vanishes to zero ($\nabla_{\theta}\mathcal{L}(\theta^{*}) = \textbf{0})$ thus we get:
    \begin{equation}
        \mathcal{L}(\theta^{*} + \Delta\theta) - \mathcal{L}(\theta^{*}) \approx \frac{1}{2} \Delta\theta^{\top} H(\theta^{*}) \Delta\theta.
    \end{equation}
    
    \begin{definition}[Parameter sensitivity]
        For a given perturbation $\Delta \theta$ of with $\|\Delta \theta\| = \epsilon$, we define \textbf{parameter sensitivity} $\text{PS}_{\mathcal{L}, \epsilon}(\theta)$ as the maximal change in the loss function:
        \begin{equation}
            \text{PS}_{\mathcal{L}, \epsilon}(\theta) = \max_{\|\Delta \theta\|=\epsilon} \left| \mathcal{L}(\theta + \Delta \theta) - \mathcal{L}(\theta) \right|.
        \end{equation}
    \end{definition}

    \begin{lemma}\label{thm:lemma-1}
        A sharp minima yields high parameter sensitivity.
    \end{lemma}
    \begin{proof}
        Since most loss functions are $C^{2}$ \cite{gao2017properties}, that is, the function, its derivatives, and its second derivatives are all continuous.
        Thus we assume the Hessian matrix to be a symmetric (interchangeable partial derivatives).
        Now consider eigen-decomposition of the Hessian $H(\theta^{*}) = U \Lambda U^\top$ where $\Lambda = \text{diag}(\lambda_1, \lambda_2, \cdots, \lambda_{p})$ and $\lambda_{\text{max}} = \lambda_1 \geq \lambda_2 \geq \cdots \geq \lambda_{p} \geq 0$ and $U$ is a orthonormal matrix consisting of eigenbasis $u_{i}$s.
        Let $\textbf{a} = U^\top \Delta\theta$ be a coordinates of $\Delta\theta$ on eigenbasis, namely
            $\Delta\theta = U\textbf{a} = \sum_{i = 1}^{p} a_{i} u_{i}$.
        Then, by substitution we get
        $\frac{1}{2} \Delta\theta^{\top} H(\theta^{*}) \Delta\theta
        =
            \frac{1}{2}\textbf{a}^{\top} \Lambda \textbf{a} 
        =
            \frac{1}{2} \sum_{i=1}^{p} \lambda_{i} a_{i}^{2}
        \leq
            \frac{1}{2}\lambda_{\text{max}} \| \textbf{a} \|^{2}
         =
            \frac{1}{2} \lambda_{\text{max}} \| \Delta\theta \|^2.$
        Thus we have
            $\text{PS}_{\mathcal{L}, \epsilon}(\theta^{*}) \approx \frac{1}{2} \text{S}_{\mathcal{L}}(\theta^{*}) \epsilon^{2}$,
        giving the desired property that the parameter sensitivity increases as the minima becomes sharper.
    \end{proof}

    By examining the change of decision values with fixed $x$, we indirectly prove that the shift of decision boundary depends on the parameter perturbation.

    \begin{lemma}\label{thm:lemma-2}
        High parameter sensitivity leads to high class decision value sensitivity.
    \end{lemma}
    \begin{proof}
        Without loss of generality, let us assume that the perturbed parameter $\theta'$ from $\theta$ is mostly related to the logit of a class $c$, namely, $z_{x; \theta'} \approx z_{x; \theta}$ and $l_{c'; \theta'}(x) \approx l_{c'; \theta}(x)$ (at least one class should be related to the parameter perturbation, and instability of the feature space is out of the current scope).
        Since the softmax-based classification loss is locally Lipschitz w.r.t. logits \cite{gao2017properties},
        \begin{equation}
            \left| \mathcal{L}(\theta') - \mathcal{L}(\theta) \right| \leq L_{c} \left| l_{c; \theta'}(x) - l_{c; \theta}(x) \right|
        \end{equation}
        holds for a Lipschitz constant $L_{c}$.
        Now consider the change of class decision values:
        \begin{equation}
            \begin{split}
                & |\text{CDV}_{c, c'}(x; \theta') - \text{CDV}_{c, c'}(x; \theta)| \\
                & = \left| \left(l_{c; \theta'}(x) - l_{c'; \theta'}(x)\right) - \left(l_{c; \theta}(x) - l_{c'; \theta}(x)\right) \right| \\
                & = \left| \left(l_{c; \theta'}(x) - l_{c; \theta}(x)\right) - \left(l_{c'; \theta'}(x) - l_{c'; \theta}(x)\right) \right| \\
                & \approx \left| l_{c; \theta'}(x) - l_{c; \theta}(x) \right| \geq \frac{1}{L_{c}} \left| \mathcal{L}(\theta') - \mathcal{L}(\theta) \right|. \\
            \end{split}
        \end{equation}
        Thus, with high parameter sensitivity and large change in the loss leads to significant shift in the decision values.
    \end{proof}

    \begin{lemma}\label{thm:lemma-3}
        Sensitive class decision boundary implies smaller class decision regions in terms of perturbation robustness.
    \end{lemma}
    \begin{proof}
        Under class decision boundary shifts of magnitude up to $\epsilon > 0$ while perturbing the parameter to $\theta'$ from $\theta$, the perturbation-safe decision region $\hat{\text{CDR}}$ shrinks as:
        \begin{equation}
            \hat{\text{CDR}}_{c}(\theta') = \{z_{x; \theta} \in \text{CDR}_{c}(\theta) \mid \text{MD}(z_{x; \theta}) > \epsilon \},
        \end{equation}
        while again, assuming that $z_{x; \theta'} \approx z_{x; \theta}$, and since $\epsilon > 0$, we get $\hat{\text{CDR}}_{c}(\theta') \subsetneq \text{CDR}_{c}(\theta)$.
    \end{proof}

    \begin{lemma}\label{thm:lemma-4}
        Smaller decision region causes higher PRS ratio.
    \end{lemma}
    \begin{proof}
        Suppose there exist two feature points $z_{x;\theta}, z_{x';\theta}$ belonging to the same populated region.
        Reducing the safe margin of class decision regions moves at least one decision boundary closer to these points, causing the decision boundary to intersect the region containing the two points.
        Geometrically, this means at least one of the feature dimensions approaches zero, flipping its sign and thus changing the feature signature for at least one point.
        Therefore, the original populated region splits into multiple populated regions, directly increasing the PRS ratio.
    \end{proof}

    \begin{theorem}
        Large-batch training is less robust to adversarial attacks.
    \end{theorem}
    \begin{proof}
        As shown in \cite{keskar2017large}, large-batch training yields sharp minima.
        By Lemma \ref{thm:lemma-1}, \ref{thm:lemma-2}, \ref{thm:lemma-3}, and \ref{thm:lemma-4}, sharp minima implies parameter sensitivity and parameter sensitivity leads to sensitive decision boundary, where sensitive decision boundary creates smaller decision regions in terms of perturbation robustness, causing higher PRS ratio, which is negatively correlated with adversarial robustness.
    \end{proof}


\subsection{Label Smoothing and PRS Ratio}

    Here, we prove explicitly that label smoothing reduces the PRS ratio.
    The proof sketch is as follows: First, we confirm that label smoothing bounds the decision value from above.
    Using this fact, we show that the rank of vectors formed by feature differences decreases, thereby demonstrating a reduction in the intra-class variance, yielding reduced PRS ratio.
    For simplicity, we redefine some concepts \cite{muller2019does}.

    \begin{definition}[Label smoothing]
        For a $K$-classed classification problem, let a data label $y$ be denote an one-hot label in $\{0, 1\}^{K}$.
        In \textbf{label smoothing}, we replace $y$ by
        \begin{equation}
            y^{(\epsilon)} = (1 - \epsilon)y + \frac{\epsilon}{K}\textbf{1},
        \end{equation}
        for class $c$ and smoothing factor $0 < \epsilon < 1$.
    \end{definition}

    \begin{lemma}\label{thm:lemma-6}
        Label smoothing bounds decision value differences from above.
    \end{lemma}
    \begin{proof}
        Note that for class $c$ and $c'$ with $c' \ne c$, we have the followings:
            $y_c^{(\epsilon)} = 1 - \epsilon + \frac{\epsilon}{K}$,
            $y_{c'}^{(\epsilon)} = \frac{\epsilon}{K}$.
        Therefore, the decision value differences at optimality are explicitly bounded:
        \begin{equation}
            \begin{split}
                & \text{CDV}_{c, c'}(x, \theta)
                     = l_{c; \theta}(x) - l_{c'; \theta}(x) \\
                    &= \log ( \exp(l_{c; \theta}(x) - l_{c'; \theta}(x)) ) \\
                    &= \log \left(
                        \frac{
                            \exp(l_{c; \theta}(x))
                            /
                            \sum_{c''}\exp(l_{c''; \theta}(x))
                        }{
                            \exp(l_{c'; \theta}(x))
                            /
                            \sum_{c''}\exp(l_{c''; \theta}(x))
                        }
                    \right) \\
                    &\Rightarrow \log \left(
                        \frac{y_{c}^{(\epsilon)}}{y_{c'}^{(\epsilon)}}
                    \right) 
                    = \log \left (
                        \frac{
                            1 - \epsilon + \frac{\epsilon}{K}
                        }{
                            \frac{\epsilon}{K}
                        }
                    \right ) \\
                    &= \log \frac{K(1 - \epsilon) + \epsilon}{\epsilon} = B < \infty.
            \end{split}
        \end{equation}
    \end{proof}

    For simplicity, let us denote $w_{cc'} = w_{c; \theta} - w_{c'; \theta}$ and $b_{cc'} = b_{c; \theta} - b_{c'; \theta}$ so that $l_{c; \theta}(x) - l_{c'; \theta}(x) = w_{cc'}z + b_{cc'}$.

    \begin{lemma}\label{thm:lemma-7}
        Bounded decision value difference reduces intra-class variance.
    \end{lemma}
    \begin{proof}
        For features $z^{(a)}, z^{(b)}$ of two samples $x^{(a)}, x^{(b)}$ from the same class $c$, since we have $\text{CDV}(c, c')(x, \theta) \rightarrow B$, we get:
        \begin{equation}
            \begin{split}
                &\text{CDV}(c, c')(x^{(a)}, \theta) - \text{CDV}(c, c')(x^{(b)}, \theta) \\
                &= (w_{cc'}z^{(a)} + b_{cc'}) - (w_{cc'}z^{(b)} + b_{cc'}) \\
                &= w_{cc'}^{\top}(z^{(a)} - z^{(b)}) 
                 \rightarrow B - B = 0, \\
            \end{split}
        \end{equation}
        for $\forall c, c'$.
        Let the matrix of class differences $W_{c}$ be
        \begin{equation}
            W_{c} = [w_{c1}, w_{c2}, \dots, w_{c\,c-1}, w_{c\,c+1}, \dots, w_{cC}] \in \mathbb{R}^{d \times (C-1)}.
        \end{equation}
        Then, the feature differences lie in the null space $\mathcal{V} = \text{Null}(W_c^\top)$.
        Namely, near optimality, the differences satisfy:
            $z^{(a)} - z^{(b)} \in \mathcal{V}$.
        Using the rank-nullity theorem, the null-space dimension satisfies:
            $\dim(\mathcal{V}) = d - \text{rank}(W_c^\top)$.
        Typically, we have $\text{rank}(W_c^\top) = K - 1 \geq 1$, and thus:
            $\dim(\mathcal{V}) = d - (K - 1) < d$.
        The intra-class covariance matrix $\Sigma_{c}$ can be represented as:
            $\Sigma_{c} = \mathbb{E}_{z \mid c}\left[
                (z - \mu_{c})(z - \mu_{c})^\top
            \right]$,
        where $\mu_{c} = \mathbb{E}_{z \mid c}\left[z\right]$.
        Note that $W_{c}^{\top}(z - \mu_{c}) = 0$ for all $z$ in class $c$.
        Since covariance matrix is always symmetric, we apply eigendecomposition to the covariance matrix as follows: $\Sigma_{y} = U\Lambda^{\top}U$ where $U = [u_{1}, u_{2}, ..., u_{d}]$ and $\Lambda = \text{diag}(\lambda_1, \lambda_2, ..., \lambda_{d})$ with $\lambda_{1} \ge \lambda_{2} \geq \cdots \ge \lambda_{d} \ge 0$.
        Note that the range of $\Sigma_{c}$ is always linear combination of $(z - \mu_{c})(z - \mu_{c})^\top v$ for some vector $v$ and expected $z$s, so by the fact that
            $W_{c}^{\top}(z - \mu_{c})(z - \mu_{c})^\top v = 0 (z - \mu_{c})^\top v = 0$,
        we have $\text{range}(\Sigma_{c}) \subset \mathcal{V}$.
        Therefore, we get $\text{rank}(\Sigma_{c}) \le \text{dim}(\mathcal{V}) = d - (K - 1)$, so at least $K - 1$ eigenvalues of $\Sigma_{c}$ must be exactly zero:
            $\lambda_{d-(K-2)} = \lambda_{d-(K-3)} = \cdots = \lambda_{d} = 0$.
        Therefore, the trace of covariance is strictly bounded by the reduced dimensionality:
            $\text{Tr}(\Sigma_{c})
            = \sum_{i=1}^{d} \lambda_{i}
            = \sum_{i=1}^{d-(K-1)} \lambda_{i}$.
        Without label smoothing however, no eigenvalues would necessarily vanish, potentially yielding larger variance:
            $\text{Tr}(\Sigma_{c}^{(\epsilon=0)})
            = \sum_{i=1}^{d} \lambda_{i}^{(\epsilon=0)}
            \ge \sum_{i=1}^{d-(K-1)} \lambda_{i}$,
        assuming that $\lambda_{i}^{(\epsilon=0)} \approx \lambda_{i}$ for $1 \le i \le d - (K - 1)$.
    \end{proof}

    \begin{theorem}
        Label smoothing lowers PRS ratio.
    \end{theorem}
    \begin{proof}
        From Lemma \ref{thm:lemma-6}, \ref{thm:lemma-7}, we get lower $\text{Var}(z | c)$, inducing higher chance of $P(\text{PR}(z^{(a)})=\text{PR}(z^{(b)}))$, where $\text{PR}$ denotes the occupying populated region, at least giving larger the number of the same signatures, thus lowering the PRS ratio.
    \end{proof}


\section{Conclusion}
\label{sec:conclusion}


    We analyze the geometrical properties of DNNs affecting adversarial robustness and introduce the Populated Region Set (PRS) to establish this relationship.
    Experiments show that PRS correlates with robustness:
    (1) Networks with low PRS ratios are more robust to gradient-based attacks and exhibit higher parameter parallelism in the final layer.
    (2) Low-PRS networks include more test samples within training regions; these included samples show higher robustness.
    (3) A PRS regularizer improves robustness without adversarial examples, and adversarial training further reduces PRS ratio while enhancing robust accuracy.
    Our work provides a geometrical interpretation of robustness through decision regions, and we expect PRS to contribute to improving DNN robustness.


\section*{Acknowledgment}

    This work was made in collaboration with Korea Advanced Institute of Science \& Technology and INEEJI Co., Ltd. Also, this work was supported by Institute of Information \& communications Technology Planning \& Evaluation (IITP) grant funded by the Korea government(MSIT) [No.RS-2022-II220984, Development of Artificial Intelligence Technology for Personalized Plug-and-Play Explanation and Verification of Explanation].



\bibliographystyle{IEEEtran}
\bibliography{references}


\end{document}